# The Impact of Image Resolution on Face Detection: A Comparative Analysis of MTCNN, YOLOv XI and YOLOv XII models

1st Ahmet Can Ömercikoğlu
*Duzce University*
*Department of Computer Engineering*
Düzce, Türkiye
ac.omercikoglu@gmail.com

2nd Mustafa Mansur Yönügül
*Duzce University*
*Department of Computer Engineering*
Düzce, Türkiye
mustafayngl@gmail.com

3rd Pakize Erdoğmuş
*Duzce University*
*Department of Computer Engineering*
Düzce, Türkiye
pakizeerdogmus@duzce.edu.tr

*Abstract*—Face detection is a crucial component in many AI-driven applications such as surveillance, biometric authentication, and human-computer interaction. However, real-world conditions like low-resolution imagery present significant challenges that degrade detection performance. In this study, we systematically investigate the impact of input resolution on the accuracy and robustness of three prominent deep learning-based face detectors: YOLOv11, YOLOv12, and MTCNN. Using the WIDER FACE dataset, we conduct extensive evaluations across multiple image resolutions (160×160, 320×320, and 640×640) and assess each model's performance using metrics such as precision, recall, mAP50, mAP50–95, and inference time. Results indicate that YOLOv11 outperforms YOLOv12 and MTCNN in terms of detection accuracy, especially at higher resolutions, while YOLOv12 exhibits slightly better recall. MTCNN, although competitive in landmark localization, lags in real-time inference speed. Our findings provide actionable insights for selecting resolution-aware face detection models suitable for varying operational constraints.

## I. Introduction

Face detection is one of the most important artificial intelligence(AI) tasks for numerous real-life applications such as facial recognition for autonomous applications, emotion detection, surveillance, security and human-computer interaction. The aim of face detection is to find the location of human faces in the images or videos, regardless of variations in pose, illumination, expression, or background.

While recent advancements in deep learning have significantly improved face detection performance, the effectiveness of these models can be heavily influenced by the quality of the input image, particularly its resolution and variations stated below. Low-resolution images, which mostly occur in outdoor scenes such as football match stadiums, or some congress. In such images, the resolution of the faces vary a lot. From the front side to back side, the size of the faces decreases, so the resolution decreases also. So this variation in one image, makes detection task difficult and degrades the performance of detection algorithms.

This study investigates the impact of image resolution on the performance of face detection systems by conducting a comparative analysis of three deep learning-based models, YOLOv11, YOLOv12, and MTCNN. By evaluating the robustness of each model under various resolutions, we aim to better understand their limitations and capabilities in real-world conditions. Our findings are intended to guide researchers and practitioners in selecting or designing models that maintain reliable detection performance even under suboptimal visual conditions.

## II. Related Works

### A. Recent Advances in Face Detection Models

Recent advances in deep learning have greatly improved the performance of face detection models. RetinaFace [1], a state-of-the-art single-shot detector specifically designed for face detection tasks, provides robust localization by jointly performing face bounding-box regression, landmark detection, and pixel-wise face segmentation.

Similarly, BlazeFace [2], has been developed for mobile and real-time applications, offers high-speed detection capabilities, especially optimized for resource-constrained environments.

FaceNet [3] is another significant model primarily developed for facial recognition tasks, utilizing deep convolutional neural networks to generate compact face embeddings. Although originally designed for recognition, FaceNet embeddings have been effectively utilized as a backbone for detection models in various real-world applications.

One of the recent study YOLO-Face V2[4], uses YOLOv5 architecture and introduces a Receptive Field Enhancement (RFE) module extracting multi-scale pixel information, detecting small faces accurately.

### B. Face Detection Challenges in Low-Resolution Conditions

Despite these advancements, face detection performance significantly deteriorates in low-resolution scenarios, commonly encountered in surveillance, outdoor events, or crowded settings. The WIDER FACE dataset benchmark study [5] highlights that face detection algorithms experience considerable drops in accuracy when encountering small, heavily occluded, or low-resolution faces, indicating a critical challenge in practical scenarios. These low-resolution conditions often result in insufficient pixel information, complicating accurate localization and identification.

To mitigate these issues, several studies proposed super-resolution-based methods aimed at enhancing face detection by artificially increasing image resolution. For instance, Yu et al. [6] proposed a face super-resolution method guided explicitly by facial component heatmaps. Their approach uses intermediate facial landmarks to reconstruct detailed facial features, significantly enhancing detection and recognition performance on low-resolution face images. Such methods provide a viable path for bridging the performance gap caused by low resolutions, although they often require additional computational overhead and specialized training pipelines.

Additionally, resolution-aware face detection methods have emerged as an alternative approach. Bai et al. [7] introduced a multi-scale convolutional neural network specifically designed for detecting faces in low-resolution images. Their model leverages multiple detection scales to robustly capture faces varying greatly in size and resolution. By explicitly modeling resolution differences, their approach demonstrated improved detection performance in challenging conditions, highlighting the effectiveness of scale-adaptive frameworks for real-world applications involving diverse resolution levels.

However, dedicated detection models specifically optimized for low-resolution conditions remain relatively limited. This research addresses this gap by systematically evaluating modern face detection models, including YOLO variants and MTCNN, under varying resolution conditions, aiming to identify their robustness and practical applicability in real-world low-resolution scenarios.

## III. Material and Methods

### A. Dataset

The WIDER FACE dataset is one of the largest and most challenging face detection datasets available. It consists of 32,203 images with 393,703 labeled face bounding boxes, covering a wide range of variations in scale, pose, occlusion, expression, and background clutter. The images were selected from the WIDER dataset, grouped into 60 different event categories, and divided into training (40%), validation (10%), and testing (50%) subsets.

Bounding boxes were carefully annotated to tightly enclose the forehead, chin, and cheeks, even under partial or heavy occlusion. Extremely small faces (10 pixels or less) were flagged for exclusion during evaluation. To account for the dataset's complexity, three levels of difficulty ('Easy', 'Medium', 'Hard') were defined based on face size, pose, and occlusion [5].

### B. Dataset Preparation

In this study, we're using the WIDER Face dataset, which is known for its detailed face annotations, to improve face detection capabilities using the YOLO (You Only Look Once) detection framework. Initially, the WIDER dataset provided bounding boxes using absolute coordinates defined by the top-left corner positions, along with width and height as well as additional details about facial attributes. Additionally, WIDER annotations were originally stored in a single text file for the entire dataset. However, YOLO requires annotations in a different format: normalized coordinates of the bounding box center and dimensions relative to the image size, with individual annotation files for each image. To bridge this gap, we developed a Python script that efficiently converts the original annotations. Our script reads each annotation entry, calculates normalized bounding box values based on image dimensions, and creates separate YOLO-compatible annotation files for every image, enabling straightforward use for training object detection models. We assigned a single class ID (e.g., 0) for all faces, since our task was single-class face detection. You can download the WIDER Face dataset converted to YOLO format from Kaggle [8] to reproduce our training and evaluation pipeline.

### C. YOLOv11

YOLOv11 represents a significant advancement in real-time object detection, building upon the well-established YOLO (You Only Look Once) framework. YOLOv11 integrates improved feature extraction capabilities and optimized convolutional architectures, enhancing detection accuracy and computational efficiency[9]. Specifically, YOLOv11 employs refined multiscale detection layers, which effectively address the challenges posed by varying object sizes within images. Moreover, it incorporates training enhancements such as mosaic augmentation and adaptive anchor allocation, enabling the model to generalize better across diverse datasets and maintain robustness in practical deployment scenarios.

### D. YOLOv12

YOLOv12 pushes the boundaries further by introducing sophisticated architectural modifications designed to enhance both speed and accuracy [10]. Central to YOLOv12's improvements is its optimized transformer-based backbone, enabling more effective global context modeling and superior performance in complex visual environments. Additionally, YOLOv12 leverages progressive distillation methods and a more effective anchor-free detection mechanism, significantly reducing the computational overhead without sacrificing detection quality. As a result, YOLOv12 achieves state-of-the-art performance, particularly excelling in scenarios demanding precise real-time inference and robust handling of densely crowded or overlapping objects.

### E. MTCNN

The Multi-task Cascaded Convolutional Neural Network (MTCNN) is a highly effective face detection framework that combines robust performance with computational efficiency. MTCNN employs a cascaded three-stage architecture, consisting of Proposal Network (P-Net), Refinement Network (R-Net), and Output Network (O-Net), each progressively refining the localization and confidence of face detections[11]. This approach significantly reduces false positives and improves accuracy, especially in challenging scenarios involving varying poses, occlusions, and lighting conditions. Moreover, MTCNN simultaneously provides accurate bounding boxes and facial landmark predictions, making it particularly well-suited for downstream tasks such as face recognition, alignment, and expression analysis.

## IV. EXPERIMENT AND RESULTS

### A. Evaluation Metrics

To thoroughly assess the performance of YOLOv11, YOLOv12, and MTCNN face detection models, we utilized several widely accepted metrics. Precision, Recall, mean Average Precision (mAP50), and mean Average Precision from IoU threshold 0.5 to 0.95 (mAP50-95) provided comprehensive measures of detection accuracy. Additionally, inference time was recorded in terms of milliseconds to quantify real-time performance. All these metrics were systematically compared across different models to identify their relative strengths and weaknesses.

In addition to precision and recall, we measured the end-to-end inference time for each model on the entire WIDER FACE validation set. Before timing, each model was "warmed up" with ten initial runs to stabilize GPU kernels. We then recorded per-image latency by marking the clock immediately before the model call and only stopping it after a torch.cuda.synchronize() (for GPU executions) to capture true completion time. Finally, we report the arithmetic mean and one standard deviation over all ~3,200 images (e.g., 12.3 ms ± 1.1 ms), conveying both typical speed and its consistency under real-world operating conditions.

### B. Experiment Environment

The models were implemented and trained using Kaggle Notebook, leveraging an environment equipped with 29GB of RAM, a 4-core CPU, and an Nvidia P100 GPU. Training was accelerated by CUDA 12.3. This standardized setup ensured fairness in comparisons, allowing performance differences to be attributed primarily to the architectural distinctions of the models.

TABLE 1. MODEL TRAINING ENVIROMENT

| Environment | Setting |
|---|---|
| OS | Kaggle Notebook |
| RAM | 29GB |
| CPU | 4 Core |
| GPU | Nvidia T4 |
| GPU Acc | CUDA 12.6 |

a.

TABLE 2. HYPER PARAMETERS

| Parameter | Value |
|---|---|
| Input Size | 640/320/160 |
| Epoch | 100 |
| Batch Size | 16 |
| Optimizer | SGD |
| Learning Rate | 0.01 |
| Momentum | 0.937 |
| Weight Decay | 0.0005 |

b.

### C. Experimental Results

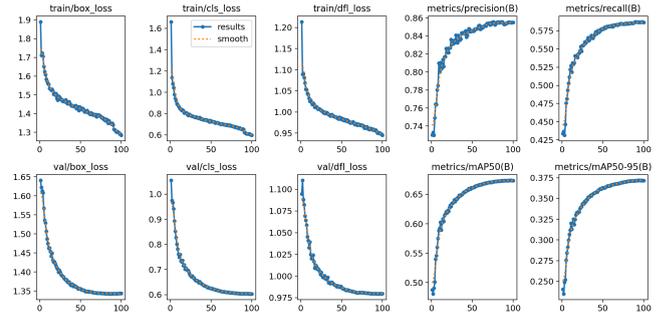

Fig. 1. YOLOv XI model training and validation performance metrics.

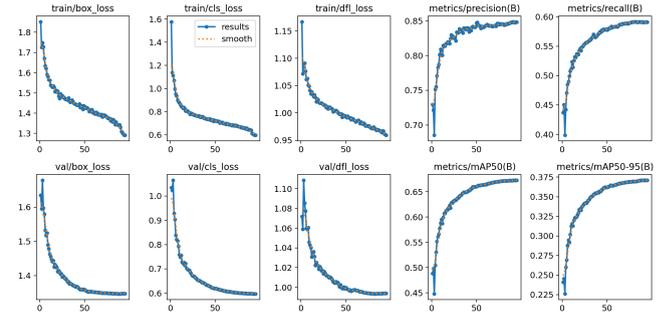

Fig. 2. YOLOv XII model training and validation performance metrics.

The experimental evaluation yielded intriguing insights regarding the face detection performance of YOLOv11, YOLOv12, and MTCNN. Based on the results summarized in the tables and visual comparisons, YOLOv11 (nano variant) demonstrated marginally better performance across most of the evaluated metrics, including Precision, mAP50, and mAP50-95, when compared to YOLOv12 (nano variant). However, YOLOv12 exhibited slightly higher Recall, indicating its capability to detect a broader range of faces, albeit with marginally lower precision. These findings highlight an interesting trend in lightweight model comparisons: YOLOv11 maintains a slight edge in accuracy over YOLOv12, suggesting that for extremely lightweight (nano-scale) configurations, YOLOv11 might be a slightly more optimal choice when prioritizing accuracy.

MTCNN struggled to detect smaller faces, resulting in lower overall mAP scores compared to both YOLO variants. This performance reflects its cascaded, multi-stage architecture, which prioritizes landmark localization and alignment accuracy over small-scale detection robustness.

To ensure a fair apples-to-apples comparison, we first converted the WIDER Face validation annotations into COCO format, as described previously: parsing the original wider_face_val_bbx_gt.txt file to extract absolute [x, y, width, height] boxes, flattening file paths to match our Kaggle directory layout, and assembling the standard COCO JSON with unique image IDs, a single "face" category, and per-annotation area fields. This one-time conversion let us leverage pycocotools.COCOeval to compute mAP@.50 and mAP@[.50–.95], as well as precision and recall at up to 100 detections, under the exact same dataset split and evaluation protocol used for our YOLO11/YOLO12 models.

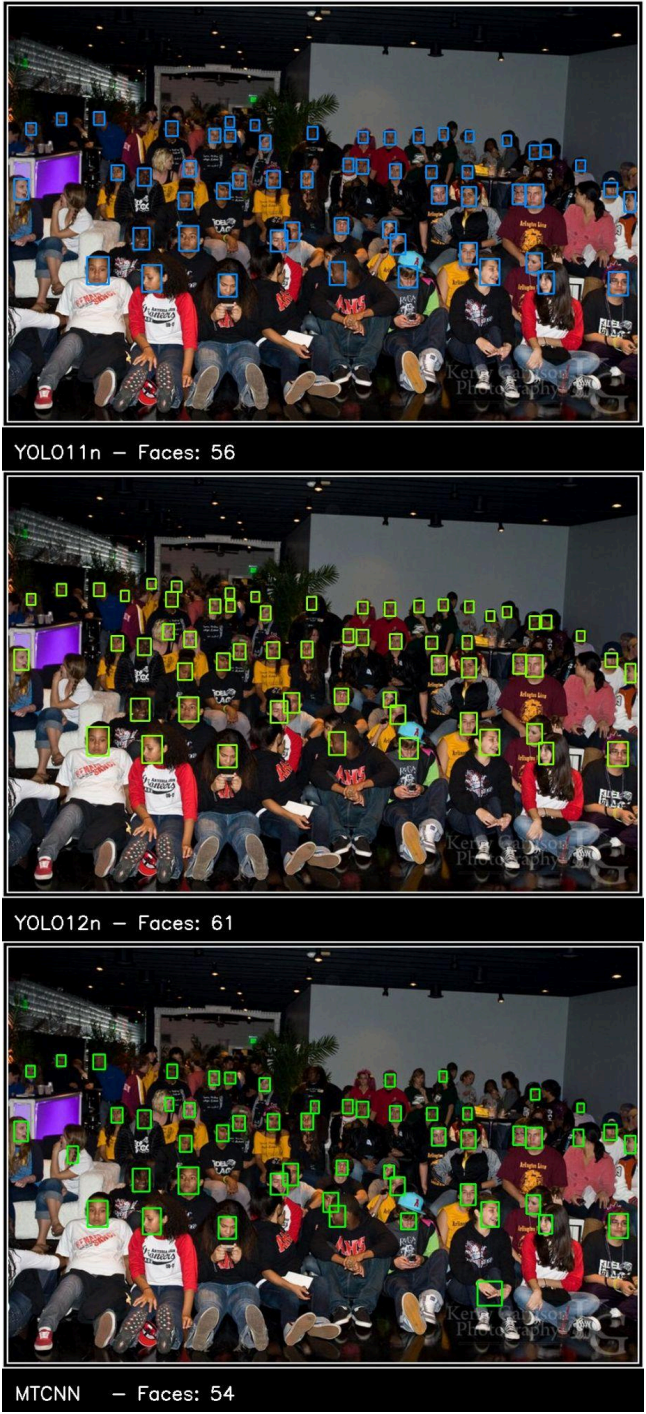

Fig. 3 Face Detection Results for YOLOv11, YOLOv12 and MTCNN.

For detection, we ran inference over all validation images using the off-the-shelf MTCNN implementation from facenet-pytorch the default pretrained model, without any fine-tuning on WIDER. We evaluated it under two settings: the library's original min_face_size=20 px threshold, and a reduced min_face_size=12 px to allow detection of smaller faces. In each case, MTCNN produced bounding boxes and confidence scores, which we reformatted into the COCO "results" JSON and passed to COCOeval. By reporting both configurations (20 px vs. 12 px) and clearly noting that this is the out-of-the-box model not trained on WIDER data, we provide a transparent baseline against which our YOLO variants' metrics can be directly compared.

TABLE 3. PERFORMANCE METRICS

| Metrics | Models | | | |
|---|---|---|---|---|
| | *YOLO v11n* | *YOLO v12n* | *MTCNN (Default)* | *MTCNN (min_face_size = 12)* |
| Precision | **85.4** | 84.8 | 39.0 | 40.0 |
| Recall | 58.7 | **59.0** | 24.4 | 24.9 |
| mAP50 | **67.3** | 67.2 | 39.0 | 40.0 |
| mAP50-95 | **37.1** | 37.0 | 21.8 | 22.4 |
| Inference Time (ms) | **29.68 ± 6.6** | 34.81 ± 2.8 | 179.84 ± 64.1 | 322.60 ± 183.02 |

In terms of inference speed and practical application potential, both YOLO models significantly outperform MTCNN. Their superior inference speeds and efficient GPU utilization highlight their appropriateness for real-time deployment scenarios such as video surveillance, interactive systems, and real-time facial analysis. Specifically, the YOLO framework's single-shot detection principle effectively balances high detection accuracy with exceptional speed, making these models particularly suitable for performance-critical use cases.

YOLOv11n achieves the fastest average inference time, closely followed by YOLOv12n, while MTCNN is notably slower. These figures represent the mean of over 3,000 sequential measurements on the full WIDER FACE validation set, with the "±" term indicating one standard deviation. To ensure a fair comparison, each model was first "warmed up" with 10 initial runs to stabilize GPU performance. Timing for each image began immediately before the model call and ended only after synchronizing the GPU, thus capturing the true end‑to‑end latency. All inferences were executed at a batch size of one and an input resolution of 640×640 px to reflect real‑time deployment conditions. Note that all timing experiments were performed in a virtualized environment, so the absolute latency values may not directly translate to real‑world deployments; however, the comparative results clearly demonstrate that the YOLO variants achieve faster and more consistent real‑time performance.

The detailed comparison of YOLOv11 and YOLOv12 across varying input resolutions (160, 320, and 640 pixels) is presented in Table X. A clear and consistent trend emerges from the results, highlighting the direct relationship between the input image resolution and the detection metrics for both YOLO models.

TABLE 4. PERFORMANCE METRICS BY IMAGE SIZE

| Models | Metrics | | | | |
|---|---|---|---|---|---|
| | *Image Size* | *Precision* | *Recall* | *mAP50* | *mAP50-95* |
| YOLO v11 | 160 | 0.620 | 0.211 | 0.245 | 0.123 |
| | 320 | 0.758 | 0.389 | 0.452 | 0.240 |
| | 640 | **0.854** | 0.587 | **0.673** | **0.371** |
| YOLO v12 | 160 | 0.614 | 0.212 | 0.245 | 0.122 |
| | 320 | 0.760 | 0.391 | 0.453 | 0.240 |
| | 640 | 0.848 | **0.590** | 0.672 | 0.370 |

Notably, as the input resolution decreases from 640 to 160 pixels, all performance metrics including Precision, Recall, mAP50, and mAP50-95 demonstrate a significant decline for both YOLOv11 and YOLOv12. For instance, YOLOv11 at an input resolution of 640 achieves a precision of 0.854, recall of 0.587, mAP50 of 0.673, and mAP50-95 of 0.371. However, at the smallest resolution of 160 pixels, these values drop substantially to 0.620, 0.211, 0.245, and 0.123 respectively. A similar pattern is evident for YOLOv12, emphasizing how lower input resolution inherently limits the model's ability to accurately detect and localize objects due to the loss of finer spatial information.

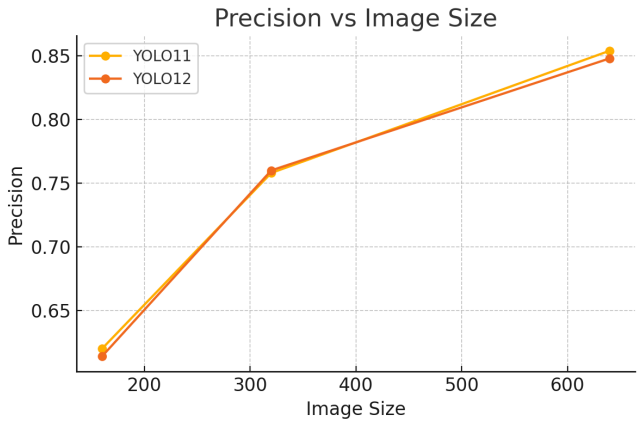

Fig. 4. Comparison.

Interestingly, YOLOv11 slightly outperforms YOLOv12 across nearly all evaluated metrics at equivalent resolutions. This subtle yet consistent advantage underscores YOLOv11's effectiveness in nano-scale scenarios, especially evident at higher resolutions (640 pixels), where it achieves marginally superior accuracy. YOLOv12, on the other hand, maintains competitive results and slightly higher recall, indicating its robust capacity to detect a greater number of objects, albeit with slightly reduced precision.

These findings suggest the critical importance of selecting appropriate input resolutions based on specific deployment constraints and accuracy requirements. Lower input resolutions, while computationally more efficient and faster in inference, inevitably sacrifice detection quality, thus requiring careful consideration depending on application priorities, such as accuracy versus computational resources.

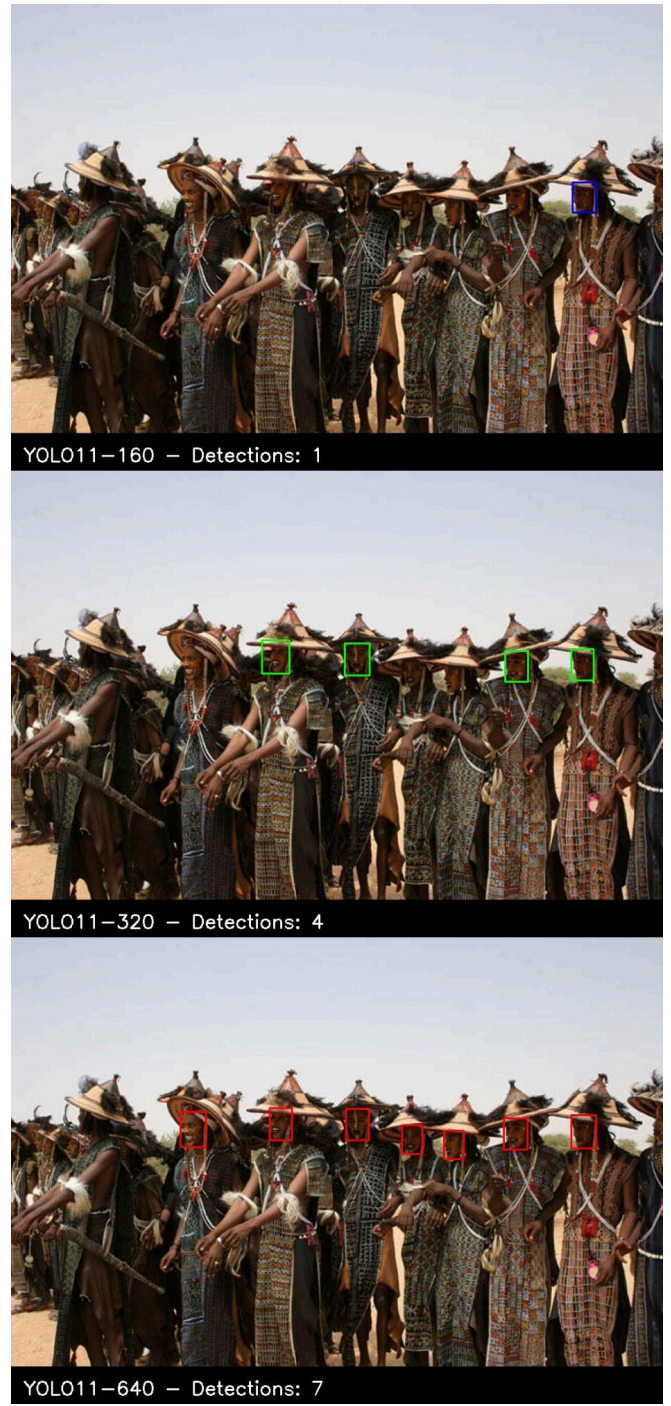

Figure 5. YOLOv11n 160v320v640 imgsz Comparison

## V. CONCLUSION

In this study, we conducted an extensive comparative evaluation of face detection models, specifically focusing on YOLOv11 (nano), YOLOv12 (nano), and MTCNN. Our experiments, structured around multiple input resolutions (160, 320, and 640 pixels), have provided insightful observations and clear guidelines for selecting appropriate models based on specific application requirements.

The experimental results demonstrate that YOLOv11 slightly outperforms YOLOv12 across most metrics, including Precision, mAP50, and mAP50-95, highlighting its effectiveness in extremely lightweight ("nano") implementations. YOLOv12, despite achieving slightly higher Recall, was generally surpassed by YOLOv11 in terms of overall accuracy metrics, particularly at higher resolutions. This nuanced performance difference suggests that YOLOv11 may represent a preferable option for tasks emphasizing detection accuracy within computationally constrained environments.

Moreover, the MTCNN model delivered suboptimal detection performance likely because it was not fine-tuned on the WIDER FACE dataset and struggles with small-scale faces and its inference speed was substantially slower than both YOLO variants. While MTCNN still provides great facial alignment and landmark localization, its combination of lower accuracy and higher latency makes it less suitable for strict real-time face detection.

The analysis of varying input resolutions clearly illustrated the trade-off between computational efficiency and detection accuracy. As image resolutions were reduced, significant performance degradation was observed across all evaluated metrics, underscoring the importance of carefully balancing speed and accuracy based on specific operational constraints.

In conclusion, the choice of face detection models and input resolutions should be guided by application-specific considerations. For real-time and resource-limited scenarios, YOLOv11 offers an optimal balance of speed and accuracy, while YOLOv12 provides marginal advantages in recall. Conversely, MTCNN remains beneficial primarily for tasks requiring detailed facial analysis rather than high-speed detection. Future work could explore hybrid strategies that combine the strengths of multiple architectures to achieve superior overall performance.